\pdfoutput=1

\documentclass[11pt]{article}

\usepackage{acl}

\usepackage{times}
\usepackage{latexsym}

\usepackage[T1]{fontenc}

\usepackage[utf8]{inputenc}

\usepackage{microtype}

\usepackage{graphicx}
\usepackage{soul}
\usepackage{url}
\usepackage{arydshln}
\usepackage{hyperref}
\usepackage{caption}
\usepackage{amsmath}
\usepackage{amsthm}
\usepackage{booktabs}
\usepackage{algorithm}
\usepackage{algorithmic}
\usepackage{multirow}
\usepackage{amsfonts}

\title{Distinguishing Non-natural from Natural Adversarial Samples for More Robust Pre-trained Language Model}

\author{Jiayi Wang, Rongzhou Bao, Zhuosheng Zhang, Hai Zhao\footnotemark[1] \\
Department of Computer Science and Engineering, Shanghai Jiao Tong University \\
Key Laboratory of Shanghai Education Commission for Intelligent Interaction\\
and Cognitive Engineering, Shanghai Jiao Tong University, Shanghai 200240, China\\
\texttt{wangjiayi\_102\_23@sjtu.edu.cn, rongzhou.bao@outlook.com}\\
\texttt{zhangzs@sjtu.edu.cn, zhaohai@cs.sjtu.edu.cn}\\
}

\begin{document}
\maketitle

\renewcommand{\thefootnote}{\fnsymbol{footnote}}
\footnotetext[1]{ Corresponding author. This work was supported in part by the Key Projects of National Natural Science Foundation of China under Grants U1836222 and 61733011.}

\renewcommand*{\thefootnote}{\arabic{footnote}}

\begin{abstract}

Recently, the problem of robustness of pre-trained language models (PrLMs) has received increasing research interest. Latest studies on adversarial attacks achieve high attack success rates against PrLMs, claiming that PrLMs are not robust. However, we find that the adversarial samples that PrLMs fail are mostly non-natural and do not appear in reality. We question the validity of current evaluation of robustness of PrLMs based on these non-natural adversarial samples and propose an anomaly detector to evaluate the robustness of PrLMs with more natural adversarial samples. We also investigate two applications of the anomaly detector: (1) In data augmentation, we employ the anomaly detector to force generating augmented data that are distinguished as non-natural, which brings larger gains to the accuracy of PrLMs. (2) We apply the anomaly detector to a defense framework to enhance the robustness of PrLMs. It can be used to defend all types of attacks and achieves higher accuracy on both adversarial samples and compliant samples than other defense frameworks. The code is available at \href{https://github.com/LilyNLP/Distinguishing-Non-Natural}{https://github.com/LilyNLP/Distinguishing-Non-Natural}.

\end{abstract}

\section{Introduction}

Pre-trained language models (PrLMs) have achieved state-of-the-art performance across a wide variety of natural language understanding tasks \cite{bert,liu2019roberta,clark2020electra}. Most works of PrLMs mainly focus on designing stronger model structures and training objectives to improve the accuracy or training efficiency. However, in real industrial applications, there exist noises that can mislead the predictions of PrLMs \cite{noise1}, which raise potential security risks and limit the application efficacy of PrLMs in practice. To solve this challenge, studies around the robustness of PrLMs have received increasing research interest. Recent studies demonstrated that, due to the lack of supervising signals and data noises in the pre-training stage, PrLMs are vulnerable to adversarial attacks, which can generate adversarial samples to fool the model \cite{survey1}. A variety of attack algorithms have been proposed to use spelling errors \cite{textbugger}, synonym substitutions \cite{TextFooler}, phrase insertions \cite{insert} or sentence structure reconstructions \cite{gan} to generate adversarial samples. Some of these attack algorithms have achieved an over 90\% attack success rate on PrLMs \cite{BertAttack, garg2020bae}. Thus they claim that existing PrLMs are not robust. 

However, we investigate the adversarial samples on which PrLMs fail, and find that most of them are not natural and fluent, thus can be distinguished by humans. These samples are unlikely to appear in reality and are against the principle that adversarial samples should be imperceptible to humans \cite{survey1}. Therefore it is not reasonable to judge the robustness of PrLMs based on these non-natural adversarial samples. By adopting a PrLM-based anomaly detector and a two-stage training strategy, we empirically demonstrate that most of the non-natural adversarial samples can be detected by the machine. Furthermore, we adopt the anomaly score (the output probability of the anomaly detector) as a constraint metric to help adversarial attacks generate more natural samples. Under this new constraint of generating natural samples, the attack success rates of existing attack methods sharply decrease. These experimental results demonstrate that the robustness of PrLMs is not as fragile as previous works claimed.

Then we explore two application scenarios of the anomaly detector. Firstly, we wonder whether the anomaly detection can generalize to other applications using artificially modified sentences. Thus we think of the data augmentation scenario. The objective of data augmentation is to increase the diversity of training data without explicitly collecting new data \cite{eda}. For an original sequence and a data augmentation technique, there exist many possible augmented sequences. We apply the anomaly detector to select among these possibilities the augmented sequence that can bring more diversity into training data. For each original sequence, we continuously generate augmented sequences until the anomaly detector distinguishes one as anomaly. The augmented data under this constraint can further increase the prediction accuracy of PrLMs than ordinary data augmentation.

Secondly, we integrate the anomaly detector into a defense framework to enhance the robustness of PrLMs. Inspired by the defense methods in the computer vision domain \cite{cv-rand1, cv-rand2, cv-rand3} which apply transformations like JPEG-based compression to mitigate the adversarial effect, we use textual transformations to restore adversarial samples. We consider a candidate set of transformation functions including back translation, MLM suggestion, synonym swap, adverb insertion, tense change, and contraction. For the input sequence that is detected as an adversarial sample, we randomly apply $k$ transformation functions from the candidate set to the sequence. We send the $k$ transformed sequences to the PrLM classifier to get their prediction scores. The final prediction is based on the average of these k prediction scores. Empirical results demonstrate that this defense framework achieves higher accuracy than other defense frameworks on both adversarial samples and compliant samples (By compliant samples, we mean the non-adversarial samples in original datasets).

\section{Related Work}

\begin{figure*}
  \centering
  \includegraphics[width =\textwidth]{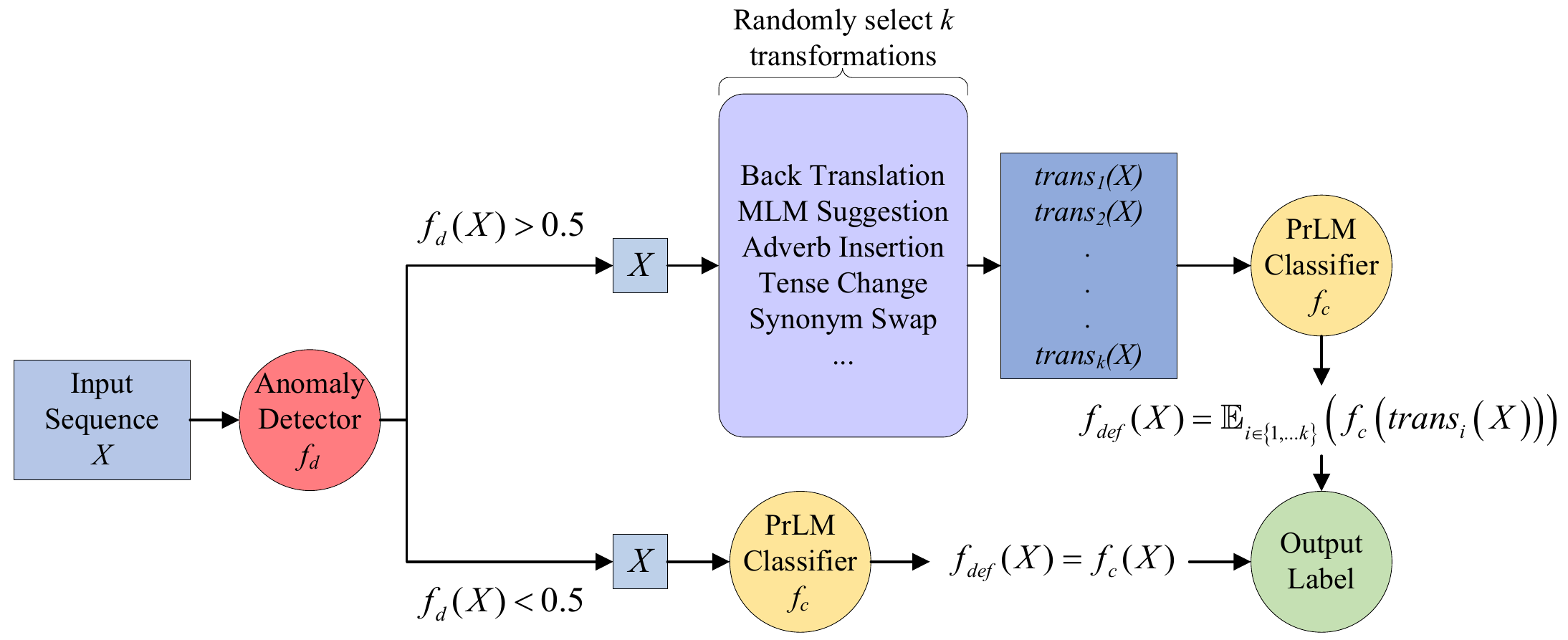}
  \caption{\label{fig:framework}Defense framework.}
\end{figure*}

The study of the robustness of PrLMs is based on the competition between adversarial attacks and defenses. Adversarial attacks find the adversarial samples where PrLMs are not robust, while defenses enhance the robustness of PrLMs by utilizing these adversarial samples or modifying model structure against the attack algorithm.

\subsection{Adversarial Attacks}
\paragraph{Problem Formulation}
Adversarial attacks generate adversarial samples against a victim model $F$, which is a PrLM-based text classifier in this paper. Given an input sequence $X$, the victim model $F$ predicts its label $F(X) = y$. The corresponding adversarial sample $X_{adv}$ should alter the prediction of the victim model and meanwhile be similar to original sequence:
\begin{equation}
\begin{aligned}
&F(X_{adv}) \neq F(X)\\
& \begin{array}{r@{\quad}l}
s.t.& d(X_{adv}, X) < \sigma,
\end{array}
\end{aligned}
\end{equation}
where $d()$ measures the size of perturbations, and $\sigma$ is a predefined threshold.

\paragraph{Classification of attacks}
Adversarial attacks can be conducted in both white-box
and black-box scenarios. In the white-box scenario \cite{white-box1}, adversarial attacks can access all information of their victim models. In the black-box scenario, adversarial attacks can only get the output of the victim models: if they get prediction scores, they are score-based attacks \cite{TextFooler}; if they get the prediction label, they are decision-based attacks \cite{decision-based1}.

According to the granularity of perturbations, textual attacks can be classified into character-level, word-level, and sentence-level attacks. Character-level attacks \cite{deepwordbug} introduce noises by replacing, inserting, or deleting a character in several words. Word-level attacks substitute several words by their synonyms to fool the model \cite{TextFooler, garg2020bae}. Sentence-level attacks generate adversarial samples by paraphrasing the original sentence \cite{scpn} or using a generative adversarial network (GAN) \cite{gan}. 

\paragraph{Metrics to constrain perturbations}
To evaluate the robustness of PrLMs, it is important that the adversarial samples are within a perturbation constraint. An adversarial sample must have similar semantic meaning to the original sample, while syntactically correct and fluent as a natural language sequence. Existing attack methods adopt the following metrics to realize this requirement:

(1) \textbf{Semantic Similarity}: semantic similarity is the most popular metric used in existing attack works \cite{TextFooler, BertAttack}. They use Universal Sentence Encoder (USE) \cite{USE} to encode original sentence and adversarial sentence into vectors and use their cosine similarity to define semantic similarity.  
    
(2) \textbf{Perturbation Rate}: perturbation rate is often used in word-level attacks \cite{TextFooler} \cite{BertAttack} to indicate the rate between the number of modified words and total words. 
    
(3) \textbf{Number of Increased Grammar Errors}: it is the number of increased grammatical errors in the adversarial sample compared to the original sample. This metric is used in \cite{HardLabel}, \cite{clare} and is calculated using LanguageTool \cite{languagetool}.
    
(4) \textbf{Levenshtein Distance}: levenshtein distance is often used in character-level attacks \cite{deepwordbug}. It refers to the number of editing operations to convert one string to another.

\subsection{Adversarial Defenses}
The objective of adversarial defenses is to design a
model which can achieve high accuracy on
both compliant and adversarial samples. One direction of adversarial defenses is adversarial training. By augmenting original training data with adversarial samples, the model is trained to be more robust to the perturbations seen in the training stage \cite{AdvTraining1}. However, it is impossible to explore all potential perturbations within a limited number of adversarial samples. Empirical results demonstrate that the improvement of robustness brought by adversarial training alone is quite limited when faced with strong dynamic attacks \cite{TextFooler, HardLabel}.

Another direction is modifying the model structure against a specific type of adversarial attack. For character-level attacks, ScRNN \cite{CombatMisspellings} leverages an RNN semi-character architecture to identify and restore the modified characters. For word-level attacks, DISP \cite{DISP} utilizes a perturbation discriminator followed by an embedding estimator to restore adversarial samples. For sentence-level attacks, DARCY \cite{DARCY} greedily searches and injects multiple trapdoors into the model to catch potential UniTrigger attacks \cite{unitrigger}. 

Certified robustness is a particular branch of defense whose aim is to ensure that the model predictions are unchanged within a perturbation scope. For example, \cite{CertifiedRobustness} and \cite{IBP1} certify the robustness of the model when input word embeddings are perturbed within the convex hull formed by the embeddings of its synonyms. However, certified robustness is hard to scale to deep networks and harms the model’s accuracy on compliant samples due to the looser outer bound.

\begin{figure*}
  \centering
  \includegraphics[width =\textwidth]{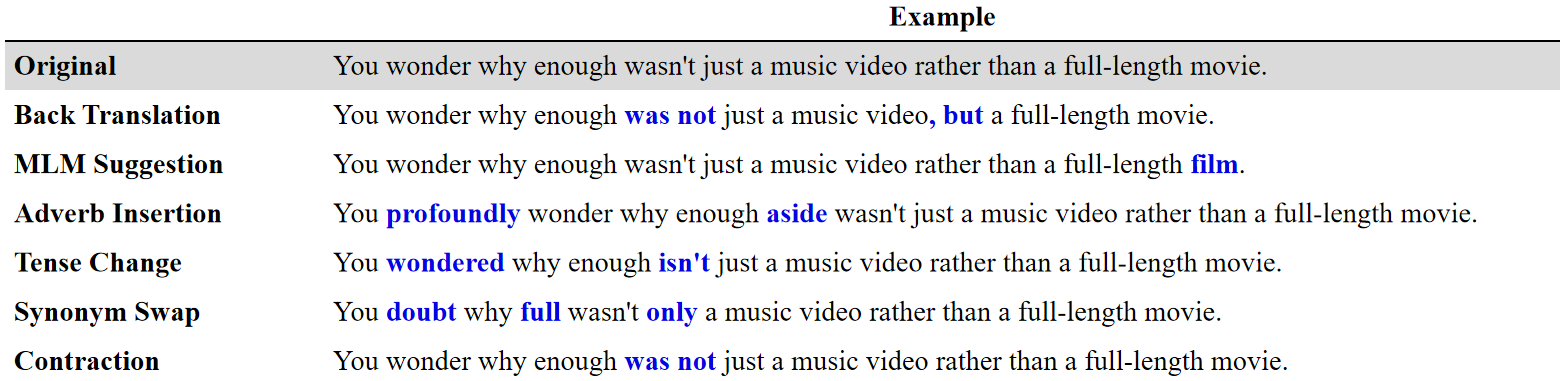}
  \caption{\label{tab:transformations}Examples of transformation functions used in the defense framework.}
\end{figure*}

\begin{figure*}
  \centering
  \includegraphics[width =\textwidth]{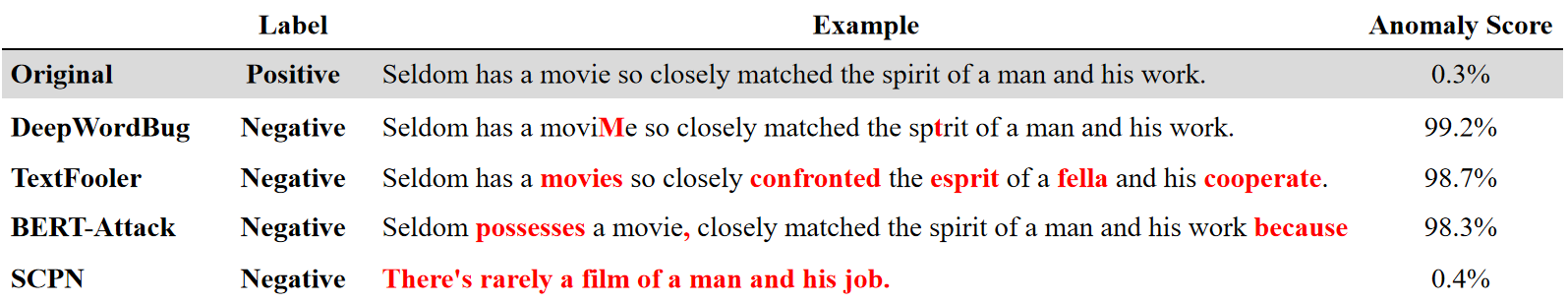}
  \caption{\label{tab:adversarial samples}Examples of adversarial samples generated by four adversarial attacks.}
\end{figure*}

\section{Methods}
\subsection{Anomaly Detector} \label{sec:3.1}
We adopt a PrLM-based binary classifier as the anomaly detector to distinguish adversarial samples from compliant samples. For an input sequence $X$, $X$ is firstly separated into sub-word tokens with a special token \texttt{[CLS]} at the beginning. A PrLM then encodes the tokens and generates a sequence of contextual embeddings $\{h_0, h_1, h_2, ..., h_n\}$, in which $h_0 \in \mathbb{R} ^H$ is the contextual representation of \texttt{[CLS]}. For text classification tasks, $h_0$ is used as the aggregate sequence representation which contains the sentence-level information. So the anomaly detector leverages $h_0$ to predict the probability that $X$ is labeled as class $\hat{y_d}$ (if $X$ is adversarial sample, $\hat{y_d} = 1$; if $X$ is compliant sample, $\hat{y_d} = 0$) by a logistic regression with softmax:
\begin{equation}
y_d = {\rm softmax}(W_d(dropout(h_0))+b_d).
\end{equation}

And we use the binary cross entropy loss function to train the anomaly detector :

\begin{equation}
loss_d = -y_d*{\rm log}
\hat{y_d}-(1-y_d)*{\rm log}
(1-\hat{y_d}).
\end{equation}

We adopt a two-stage training strategy for the anomaly detector. In the first stage, we generate the "artificial samples" using the same way each attack modifies the sentence (details of how attacks modify the sentences are described in Section \ref{sec:4.2}). But the artificial samples are not required to alter the prediction result of the PrLM, so the modification is only applied once. For example, to generate artificial samples simulating the word-level attack TextFooler, we substitute a portion of words by their synonyms in a synonym set according to WordNet. The training data consist of original samples (labeled as 0) in the train set and their corresponding artificial samples (labeled as 1). We train the detector on these data so that it can learn to distinguish artificially modified sequences from natural sequences. In the second stage, we generate the adversarial samples (labeled as 1) from the original samples (labeled as 0) in the train set, and train the anomaly detector to distinguish adversarial samples from original samples. In this way, the detector can distinguish non-naturally modified examples, and especially the adversarial ones among them. The experimental results in section \ref{sec:5.1} demonstrate that the anomaly detector can accurately distinguish adversarial samples from compliant samples.

\begin{table}[htb!]
    \centering
    \small
    \setlength{\tabcolsep}{3pt}
    {
        \begin{tabular}{llccc}
        \toprule
        Task & Dataset & Train & Test & Avg Len\\
        \midrule
        \multirow{3}{*}{Classification} & MR & 9K & 1K & 20 \\
        &SST2 & 67K & 1.8K & 20 \\
        &IMDB & 25K & 25K & 215 \\
        \midrule
        Entailment & MNLI & 433K & 10K & 11 \\
        \bottomrule
        \end{tabular}
    }
\caption{\label{tab:compairison} Dataset statistics.}
\label{tab:booktabs}
\end{table}

\begin{table*}[htb!]
	\centering
	\small
	\setlength{\tabcolsep}{3pt}
	{
        \begin{tabular}{l|ccc|ccc|ccc|ccc}
        \toprule
        
        { } &
        \multicolumn{3}{c|}{MR} &
        \multicolumn{3}{c|}{SST2} &
        \multicolumn{3}{c|}{IMDB} &
        \multicolumn{3}{c}{MNLI}\\
        &
        TPR. &
        FPR. &
        F1. &
        TPR. &
        FPR. &
        F1. &
        TPR. &
        FPR. &
        F1. &
        TPR. &
        FPR. &
        F1.\\
        \hline
        DeepWordBug & 96.2 & 1.3 & 97.4 & 98.5 & 3.7 & 97.4 & 94.4 & 1.6 & 96.3 & 97.6 & 9.2 & 94.4  \\
        TextFooler & 80.2 & 3.8 & 87.2 & 90.6 & 18.9 & 86.5 & 83.6 & 2.6 & 89.8 & 87.6 & 11.0 & 88.2 \\
        BERT-Attack & 72.6 & 4.0 & 81.9 & 86.5 & 12.8 & 87.1 & 87.2 & 3.2 & 91.6 & 86.4 & 13.0 & 86.7\\
        SCPN & 94.5 & 4.1 & 95.2 & 94.6 & 12.6 & 88.2 & - & - & - & 93.0 & 13.4 & 90.0\\
        \bottomrule
        
        \end{tabular}
    }
    \caption{\label{tab:specific detector} Performance of anomaly detector trained on each dataset and each attack method.}
\end{table*}

\begin{table*}
	\centering
	\small
	\setlength{\tabcolsep}{3pt}
	{
        \begin{tabular}{l|cc|cc|cc|cc}
        \toprule
        
        {} &
        \multicolumn{2}{c|}{MR} &
        \multicolumn{2}{c|}{SST2} &
        \multicolumn{2}{c|}{IMDB} &
        \multicolumn{2}{c}{MNLI} \\
        &
        w/o Cons. &
        w Cons. &
        w/o Cons. &
        w Cons. &
        w/o Cons. &
        w Cons. &
        w/o Cons. &
        w Cons. \\
        \hline
        Deepwordbug & 82.2 & 8.5 & 78.3 & 2.8  & 74.2 & 23.2 & 76.8 & 25.2\\
        TextFooler & 80.5 & 35.2 & 61.0 & 31.4 & 86.6 & 40.4 & 86.5 & 38.3 \\
        BERT-Attack & 84.7 & 13.9 & 87.2 & 11.5 & 87.5 & 18.9 & 89.8 & 15.2 \\
        \bottomrule
        
        \end{tabular}
    }
    \caption{\label{tab:constraint} The attack success rate of attacks using BERT as victim model without and with the anomaly score constraint on MR, SST2, IMDB, MNLI.}
\end{table*}

\begin{table*}
	\centering
	\small
	\setlength{\tabcolsep}{3pt}
	{
        \begin{tabular}{l|cc|cc|cc}
        \toprule
        
        {} &
        \multicolumn{2}{c|}{BERT} &
        \multicolumn{2}{c|}{RoBERTa} &
        \multicolumn{2}{c}{ELECTRA} \\
        &
        w/o Cons. &
        w Cons. &
        w/o Cons. &
        w Cons. &
        w/o Cons. &
        w Cons. \\
        \hline
        Deepwordbug & 82.2 & 8.5 & 83.8 & 10.4 & 79.4 & 7.9 \\
        TextFooler & 80.5 & 35.2 & 67.6 & 36.3 & 63.6 & 33.6 \\
        BERT-Attack & 84.7 & 13.9 & 73.7 & 17.4 & 70.8 & 14.2 \\
        \bottomrule
        
        \end{tabular}
    }
    \caption{\label{tab:prlms} The attack success rate of attacks without and with the anomaly score constraint using different PrLMs as victim models on MR.}
\end{table*}

\begin{table}
	\centering
	\small
	\setlength{\tabcolsep}{3pt}
	{
        \begin{tabular}{l|ccc}
        \toprule
        & No &
        Augmentation &
        Augmentation
        \\
        &
        Augmentation &
        w/o Selection &
        w Selection\\
        \hline
        BERT  & 86.4 & 87.1 & 88.3  \\
        RoBERTa & 88.3 & 89.1 & 89.5 \\
        ELECTRA & 90.1 & 90.2 & 90.4 \\
        \bottomrule
        
        \end{tabular}
    }
    \caption{\label{tab:data augmentation} The accuracy of no augmentation, after the data augmentation without and with the selection of detector on MR.}
\end{table}

\begin{table*}
	\centering
	\small
	\setlength{\tabcolsep}{3pt}
	{
        \begin{tabular}{l|cc|cc|cc|cc}
        \toprule
        
        {} &
        \multicolumn{2}{c|}{MR} &
        \multicolumn{2}{c|}{SST2} &
        \multicolumn{2}{c|}{IMDB} &
        \multicolumn{2}{c}{MNLI} \\
        &
        w/o Def.&
        w Def.&
        w/o Def.&
        w Def.&
        w/o Def.&
        w Def.&
        w/o Def.&
        w Def. \\
        \hline
        DeepWordBug & 16.3 & 57.5 & 19.7 & 62.3 & 24.3 & 81.4 & 18.7 & 70.3\\
        TextFooler & 16.7 & 66.8 & 36.2 & 73.3 & 12.4 & 90.3 & 11.3 & 69.2\\
        BERT-Attack & 13.3 & 61.5 & 12.8 & 65.2 & 11.8 & 85.9 & 9.5 & 65.4 \\
        SCPN & 64.2 & 74.3 & 70.8 & 81.5 & - & - & 66.9 & 75.0  \\
        \bottomrule
        \end{tabular}
        \caption{\label{tab:adversarial accuracy} The adversarial accuracy with and without defense using BERT as victim model.}
    }
\end{table*}

\begin{table}
	\centering
	\small
	\setlength{\tabcolsep}{3pt}
	{
        \begin{tabular}{lcccc}
        \toprule
        {} &
        MR &
        SST2 &
        IMDB &
        MNLI\\
        \midrule
        w/o Def. & 86.4 & 92.6 & 92.4 & 84.0 \\
        w Def. & 87.0 & 92.6 & 92.5 & 84.0 \\
        \bottomrule
        \end{tabular}
        \caption{\label{tab:original accuracy} The original accuracy with and without defense using BERT as victim model.}
    }
\end{table}

\begin{table*}
	\centering
	\small
	\setlength{\tabcolsep}{3pt}
	
	{
        \begin{tabular}{l|cc|cc|cc|cc}
        \toprule
        
        {} &
        \multicolumn{2}{c|}{MR} &
        \multicolumn{2}{c|}{SST2} &
        \multicolumn{2}{c|}{IMDB} &
        \multicolumn{2}{c}{MNLI} \\
        &
        Orig\%&
        Adv\%&
        Orig\%&
        Adv\%&
        Orig\%&
        Adv\%&
        Orig\%&
        Adv\% \\
        \hline
        No Defense & 86.4 & 16.7 & 92.6 & 36.2 & 92.4 & 12.4 & 84.0 & 11.3\\
        \hdashline
        Adv Training & 85.4 & 35.2 & 92.1 & 48.5 & 92.2 & 34.3 & 82.3 & 33.5\\
        DISP & 82.0 & 42.1 & 91.1 & 69.8 & 91.7 & 81.9 & 76.3 &  35.2\\
        SAFER & 79.0 & 55.3 & 90.8 & \textbf{75.1} & 91.3 & 88.1 & 82.1  & 54.7\\
        Ours & \textbf{87.0} & \textbf{66.8} & \textbf{92.6} & 73.3 & \textbf{92.5} & \textbf{90.3} & \textbf{84.0} & \textbf{69.2}  \\
        \bottomrule
        \end{tabular}
        \caption{\label{tab:comparison} The performance of our defense framework compared with other word-level defenses using BERT as PrLM and TextFooler as attack. Orig\% is the original accuracy and Adv\% is the adversarial accuracy.}
    }
\end{table*}

\subsection{Evaluation of Robustness under Anomaly Score Constraint}
Existing adversarial samples have applied some thresholds to limit the anomaly of adversarial samples. However, the generated adversarial samples are still not natural, indicating that existing metrics are not effective enough. In order to measure the robustness of PrLMs with more natural adversarial samples, we use a new metric: anomaly score, to constrain the perturbations. Given a sentence $X$, we leverage the probability that $X$ is adversarial sample predicted by anomaly detector as the anomaly score of $X$:  
\begin{equation}
Score(X) = Prob(\hat{y_d} = 1|X).
\end{equation}

For existing attacks, we add a threshold on anomaly score to enforce the attacks to generate more natural and undetectable adversarial samples. The attack problem formulation now becomes:
\begin{equation}
\begin{aligned}
&F(X_{adv}) \neq F(X)\\
& \begin{array}{r@{\quad}l}
s.t.& d(X_{adv}, X) < \sigma, \\
& Score(X_{adv}) < 0.5, \\
\end{array}
\end{aligned}
\end{equation}
where $d()$ measures the perceptual difference between $X_{adv}$ and $X$. Each attack has its own definition of $d()$ and threshold $\sigma$. And we add on a new constraint that the anomaly score of $X_{adv}$ should be smaller than 0.5. We investigate the robustness of PrLMs under the constraint of anomaly score and find that PrLMs are more robust than previously claimed.

\subsection{Application in Data Augmentation}
In data augmentation, PrLM is trained on original sentences and their artificially augmented sentences to improve the diversity of training data. We consider random synonym substitution as the augmentation technique for experiments. For an original sequence of $n$ words, we randomly select $p\% * n$ words and substitute them with their synonyms to form the augmented sequence. For each replaced word, the replacing synonym is randomly selected among its $s$ most similar synonyms. So we will have in total $C_n^{p\%*n}*s^{p\%*n}$ possible augmented sequences. In order to select the augmented sequence that can bring more diversity into training data, we apply the anomaly detector to select the augmented sequence that is distinguished as anomaly. For each original sequence, we continuously apply random synonym substitution to form candidate augmented sequences until the detector distinguishes one as anomaly.

\subsection{Application in Enhancing Robustness}

There are two ways to apply the anomaly detector in enhancing the robustness of PrLMs: (1) detect and then directly block the adversarial samples; (2) distinguish the adversarial samples and conduct operations on them to make the PrLMs give correct predictions. The first application is trivial so we explore the second way. 

We propose a defense framework as shown in Figure \ref{fig:framework}. We firstly build a transformation function set containing $t$ transformation function candidates: \textit{Back Translation} (translate the original sentence into another language and translate it back to original language); \textit{MLM Suggestion} (mask several tokens in the original sentence and use masked language model to predict the masked tokens); \textit{Adverb Insertion} (insert adverbs before verbs); \textit{Tense Change} (change the tense of verbs into another tense); \textit{Synonym Swap} (swap several words with their synonyms according to WordNet), \textit{Contraction} (contract or extend the original sentence by common abbreviations). We implement these transformation functions based on \cite{textflint} \footnote{https://github.com/textflint/textflint}. The examples of these transformation functions are displayed in Figure \ref{tab:transformations}.

For each input sequence $X$, we apply the anomaly detector $f_d$ to identify whether it is adversarial ($f_d(X) > 0.5$) or not ($f_d(X) < 0.5$). If the $X$ is recognized as compliant sample, it will be directly sent to the PrLM classifier $f_c$ to get the final output probability of the defense framework: $f_{def}(X) = f_c(X)$. If $X$ is recognized as adversarial sample, we will randomly select $k$ transformation functions from the transformation candidate set and apply them to $X$. We send the $k$ transformed sequences $trans_i(X), i \in \{1,...,k\}$ to the PrLM classifier to get their prediction probabilities $f_c(trans_i(X)), i \in \{1,...,k\}$, and the final prediction probability of the defense framework is the expectation over the $k$ transformed probabilities $f_{def}(X) = \mathbb{E}_{i \in \{1,...,k\}}(f_c(trans_i(X)))$.

Since the detector is not perfect, there always exist a small number of compliant samples that are misclassified into adversarial samples. In order to minimize the harm to the accuracy of PrLMs on compliant samples, during the training stage of PrLMs, we augment the training data with their transformed data. In this way, the PrLMs are more stable to transformations on compliant samples, and data augmentation itself also brings gains to the accuracy of PrLMs.

\section{Experimental Implementation}
\subsection{PrLMs}
We investigate three PrLMs: BERT$_\text{BASE}$  \cite{bert}, RoBERTa$_\text{BASE}$ \cite{liu2019roberta} and ELECTRA$_\text{BASE}$ \cite{clark2020electra}. The PrLMs are all implemented in their base-uncased version based on PyTorch \footnote{https://github.com/huggingface}: they each have 12 layers, 768 hidden units, 12 heads and around 100M parameters. For most experiments on attacks and defenses, we use BERT$_\text{BASE}$ as the victim model for an easy comparison between our results and those of previous works. 

\subsection{Adversarial Attacks} \label{sec:4.2}

We investigate four adversarial attacks from character level, word level to sentence level. Examples of adversarial samples generated by these four attacks are demonstrated in Figure \ref{tab:adversarial samples}.

\paragraph{Character-level attack}
For character-level attack, we consider Deepwordbug, which applies four types of character-level modifications (substitution, insertion, deletion and swap) to words in the original sample. Edit distance is used to constrain the similarity between original and adversarial sentences.

\paragraph{Word-level attack}
We select two classic word-level attack methods: TextFooler \cite{TextFooler} and BERT-Attack \cite{BertAttack}. They both sort the words in the original sample by importance scores, and then substitute the words in order with their synonyms until the PrLM is fooled. TextFooler selects the substitution word from a synonym set of the original word according to WordNet \cite{CounterFittingWordVectors}. BERT-Attack masks the original word and uses the masked language model (MLM) to predict the substitution word. Semantic similarity and perturbation rate are used to constrain the perturbation size.

\paragraph{Sentence-level attack}
We select SCPN \footnote{https://github.com/thunlp/OpenAttack} \cite{scpn} to generate sentence-level adversarial samples. SCPN applies syntactic transformations to original sentences and automatically labels the sentences with their syntactic transformations. Based on these labeled data, SCPN trains a neural encoder-decoder model to generate syntactically controlled paraphrased adversarial samples. Semantic similarity is used to ensure that the semantic meaning remains unchanged.

\subsection{Datasets}

Experiments are conducted on four datasets: \textbf{SST2} \cite{SST2}, \textbf{MR} \cite{mr}, \textbf{IMDB} \cite{IMDB}, \textbf{MNLI} \cite{MNLI}, covering two major NLP tasks: text classification and natural language inference (NLI). The dataset statistics are displayed in Table \ref{tab:compairison}. 

For text classification task, we use three datasets with average text lengths from 20 to 215 words in English: (1) \textbf{SST2} \cite{SST2}: a phrase-level binary sentiment classification dataset on movie reviews; (2)  \textbf{MR} \cite{mr}: a sentence-level binary sentiment classification dataset on movie reviews; (3)  \textbf{IMDB} \cite{IMDB} : a document-level binary sentiment classification dataset on movie reviews.  For the NLI task, we use \textbf{MNLI} \cite{MNLI}, a widely adopted NLI benchmark with coverage of the transcribed speech, popular fiction, and government reports. When attacking the NLI task, we keep the original premises unchanged and generate adversarial hypotheses.

\subsection{Experimental Setup}
The hyperparameter $k$ in the defense framework is 3. For the victim PrLMs under attack, we fine-tune PrLMs on the training set of each dataset. For the anomaly detector, we use BERT$_\text{BASE}$ as the base PrLM and fine-tune it on the training data indicated in Section \ref{sec:3.1}. For the data augmentation, we fine-tune PrLMs on the augmented training set of each dataset. During the fine-tuning of all these PrLMs, we use AdamW \cite{DBLP:journals/corr/abs-1711-05101} as our optimizer with a learning rate of 3e-5 and a batch size of 16. The number of training epochs is set to 5. To avoid randomness, we report the results of applications in data augmentation and defense framework based on the
average of 3 runs.

\section{Experimental Results}
\subsection{Anomaly Detector} \label{sec:5.1}

We consider three metrics to evaluate the performance of the anomaly detector: \textit{F1 score} (F1); \textit{True Positive Rate} (TPR): the percentage of adversarial samples that are correctly identified; \textit{False Positive Rate} (FPR): the percentage of compliant samples that are misidentified as adversarial. The experimental results are shown in Table \ref{tab:specific detector}. The results of SCPN on the IMDB dataset are unavailable since SCPN cannot tackle document-level texts. Empirical results demonstrate that the anomaly detector can achieve an average F1 score over 90\%, an average TPR over 88\%, and an average FPR less than 10\% for adversarial attacks from character-level, word-level to sentence-level. 

\subsection{Evaluation of Robustness under Anomaly Score Constraint}

We now conduct different types of attacks under the constraint that the anomaly score of generated adversarial samples should be less than 0.5. Table \ref{tab:constraint} compares the attack success rate of different attacks with and without the anomaly score constraint when the victim PrLM is BERT. We can observe a sharp decrease in attack success rate with the new constraint for all levels of attacks. This result is surprising in that the attackers examined are dynamic. Despite their iterative attempts to attack the model, the attackers fail to generate a natural adversarial sample that can bypass the anomaly detector.

To ensure that this phenomenon holds for other PrLMs, we conduct experiments on RoBERTa and ELECTRA. As shown in Table \ref{tab:prlms}, the attack success rates also drop markedly under the constraint of anomaly score for these PrLMs. These empirical results demonstrate that PrLMs are more robust than previous attack methods have claimed, given that most of the adversarial samples generated by previous attacks are non-natural and detectable. However, there still exist a little portion of undetectable adversarial samples that can successfully mislead PrLMs.

\subsection{Application in Data Augmentation}
We consider the random synonym substitution that substitutes 30\% words with their synonyms selected in 50 most similar words. Table \ref{tab:data augmentation} compares the accuracy after data augmentation without and with the selection of anomaly detector. We can observe a further increase in accuracy with the selection of the anomaly detector. However, the stronger the PrLM is, the smaller the increase is.

\subsection{Application in Enhancing Robustness of PrLMs}

We evaluate the performance of the defense framework based on original accuracy and adversarial accuracy. The original accuracy is the prediction accuracy of the defense framework on original compliant samples. The adversarial accuracy is the accuracy of the defense framework after the attack. Here we consider the situation that the attack algorithm can iteratively generate adversarial samples against our defense framework until it succeeds or exceeds the upper limit of attempts.

Table \ref{tab:adversarial accuracy} shows the adversarial accuracy with and without the defense using BERT as the victim PrLM. We can see a large improvement in the adversarial accuracy with the defense for all levels of attacks. Table \ref{tab:original accuracy} shows the original accuracy with and without the defense. We find that the original accuracy gets even higher with the defense. This is because with anomaly detection, the transformations are only applied to detected adversarial examples. For the very few compliant sentences that are detected by mistake as anomaly and then applied transformations, the data augmentation in the training stage has trained the PrLMs to be stable to transformations on compliant samples. Besides, the data augmentation alone brings an increase to the original accuracy. Therefore the proposed framework does not harm and even increases the prediction accuracy for non-adversarial samples, which is important in real application scenarios.

Since word-level attacks are the most influential and widely-used type of attack, we compare the performance of our defense framework with several state-of-the-art word-level defenses (adversarial training, DISP, SAFER) while facing TextFooler as the attack model. DISP \cite{DISP} detects and restores adversarial examples by leveraging a perturbation discriminator and an embedding estimator. SAFER \cite{safer} smooths the classifier by averaging the outputs of a set of randomized examples. As shown in Table \ref{tab:comparison}, although DISP and SAFER are especially designed for word-level attacks, our defense framework outperforms them in most cases on both original accuracy and adversarial accuracy.

\section{Discussion}

There are two trade-offs for the defense framework:

(1) The trade-off between original accuracy and adversarial accuracy. If we abandon the anomaly detector and apply random transformations to all input sequences, then the adversarial accuracy can further increase by 5-7\%, but the original accuracy will decrease by 1-3\%. Since in real applications it is not reasonable to sacrifice too much precision for possible security problems, we adopt the anomaly detector to preserve the original accuracy. However, by developing a stronger detector with a higher TPR, the defense framework has the potential to achieve higher adversarial accuracy. 

(2) The trade-off between training efficiency and original accuracy. To preserve the original accuracy, we apply data augmentation in the training stage of the defense framework to make it more stable to transformations on compliant samples. However, the training cost is now multiplied by the size of the transformation set $n$ ($n=6$ in the experimental realization). If we abandon the data augmentation in training stage, the training efficiency of the defense framework is the same as the vanilla fine-tuning of PrLM, but the original accuracy will decrease by 0.5-1.5\%. 

A limitation of our work is that the attacks we examined are black-box or grey-box attacks, but do not include white-box (gradient-based) attacks. However, since more than 75\% of the existing textual attacks are not based on gradient \footnote{https://github.com/textflint/textflint}, the defense framework is effective for the majority of attacks. We will investigate white-box attacks in future works.

\section{Conclusion}
In this study, we question the validity of the current evaluation of robustness of PrLMs based on non-natural adversarial samples, and propose an anomaly detector to evaluate the robustness of PrLMs with more natural adversarial samples. To increase the precision of PrLMs, we employ the anomaly detector to select the augmented data that are distinguished as anomaly to introduce more diversity in the training stage. The data augmentation after selection brings larger gains to the accuracy of PrLMs. To enhance the robustness of PrLMs, we integrate the anomaly detector to a defense framework using expectation over randomly selected transformations. This defense framework can be used to defend all levels of attacks, while achieving higher accuracy on both adversarial samples and compliant samples than other defense frameworks targeting specific levels of attack.

\bibliography{acl_latex}

\begin{thebibliography}{38}
\expandafter\ifx\csname natexlab\endcsname\relax\def\natexlab#1{#1}\fi

\bibitem[{Cer et~al.(2018)Cer, Yang, Kong, Hua, Limtiaco, St.~John, Constant,
  Guajardo-Cespedes, Yuan, Tar, Strope, and Kurzweil}]{USE}
Daniel Cer, Yinfei Yang, Sheng-yi Kong, Nan Hua, Nicole Limtiaco, Rhomni
  St.~John, Noah Constant, Mario Guajardo-Cespedes, Steve Yuan, Chris Tar,
  Brian Strope, and Ray Kurzweil. 2018.
\newblock \href {https://doi.org/10.18653/v1/D18-2029} {Universal sentence
  encoder for {E}nglish}.
\newblock In \emph{Proceedings of the 2018 Conference on Empirical Methods in
  Natural Language Processing: System Demonstrations}, pages 169--174,
  Brussels, Belgium. Association for Computational Linguistics.

\bibitem[{Clark et~al.(2020)Clark, Luong, Le, and Manning}]{clark2020electra}
Kevin Clark, Minh-Thang Luong, Quoc~V. Le, and Christopher~D. Manning. 2020.
\newblock {ELECTRA}: Pre-training text encoders as discriminators rather than
  generators.
\newblock In \emph{ICLR}.

\bibitem[{Das et~al.(2017)Das, Shanbhogue, Chen, Hohman, Chen, Kounavis, and
  Chau}]{cv-rand2}
Nilaksh Das, Madhuri Shanbhogue, Shang-Tse Chen, Fred Hohman, Li~Chen,
  Michael~E. Kounavis, and Duen~Horng Chau. 2017.
\newblock \href {http://arxiv.org/abs/1705.02900} {Keeping the bad guys out:
  Protecting and vaccinating deep learning with jpeg compression}.

\bibitem[{Devlin et~al.(2018)Devlin, Chang, Lee, and Toutanova}]{bert}
Jacob Devlin, Ming-Wei Chang, Kenton Lee, and Kristina Toutanova. 2018.
\newblock {BERT}: Pre-training of deep bidirectional transformers for language
  understanding.
\newblock \emph{arXiv preprint arXiv:1810.04805}.

\bibitem[{Gao et~al.(2018)Gao, Lanchantin, Soffa, and Qi}]{deepwordbug}
Ji~Gao, Jack Lanchantin, Mary~Lou Soffa, and Yanjun Qi. 2018.
\newblock \href {https://doi.org/10.1109/SPW.2018.00016} {Black-box generation
  of adversarial text sequences to evade deep learning classifiers}.
\newblock In \emph{2018 IEEE Security and Privacy Workshops (SPW)}, pages
  50--56.

\bibitem[{Garg and Ramakrishnan(2020)}]{garg2020bae}
Siddhant Garg and Goutham Ramakrishnan. 2020.
\newblock \href {http://arxiv.org/abs/2004.01970} {Bae: Bert-based adversarial
  examples for text classification}.

\bibitem[{Goodfellow et~al.(2015)Goodfellow, Shlens, and
  Szegedy}]{AdvTraining1}
Ian~J. Goodfellow, Jonathon Shlens, and Christian Szegedy. 2015.
\newblock Explaining and harnessing adversarial examples.

\bibitem[{Huang et~al.(2019)Huang, Stanforth, Welbl, Dyer, Yogatama, Gowal,
  Dvijotham, and Kohli}]{IBP1}
Po-Sen Huang, Robert Stanforth, Johannes Welbl, Chris Dyer, Dani Yogatama, Sven
  Gowal, Krishnamurthy Dvijotham, and Pushmeet Kohli. 2019.
\newblock \href {https://doi.org/10.18653/v1/D19-1419} {Achieving verified
  robustness to symbol substitutions via interval bound propagation}.
\newblock In \emph{Proceedings of the 2019 Conference on Empirical Methods in
  Natural Language Processing and the 9th International Joint Conference on
  Natural Language Processing (EMNLP-IJCNLP)}, pages 4083--4093, Hong Kong,
  China. Association for Computational Linguistics.

\bibitem[{Iyyer et~al.(2018)Iyyer, Wieting, Gimpel, and Zettlemoyer}]{scpn}
Mohit Iyyer, John Wieting, Kevin Gimpel, and Luke Zettlemoyer. 2018.
\newblock \href {http://arxiv.org/abs/1804.06059} {Adversarial example
  generation with syntactically controlled paraphrase networks}.

\bibitem[{Jia et~al.(2019)Jia, Raghunathan, Göksel, and
  Liang}]{CertifiedRobustness}
Robin Jia, Aditi Raghunathan, Kerem Göksel, and Percy Liang. 2019.
\newblock \href {http://arxiv.org/abs/1909.00986} {Certified robustness to
  adversarial word substitutions}.

\bibitem[{Jin et~al.(2020)Jin, Jin, Zhou, and Szolovits}]{TextFooler}
Di~Jin, Zhijing Jin, Joey~Tianyi Zhou, and Peter Szolovits. 2020.
\newblock \href {https://doi.org/10.1609/aaai.v34i05.6311} {Is bert really
  robust? a strong baseline for natural language attack on text classification
  and entailment}.
\newblock \emph{Proceedings of the AAAI Conference on Artificial Intelligence},
  34(05):8018--8025.

\bibitem[{Le et~al.(2021)Le, Park, and Lee}]{DARCY}
Thai Le, Noseong Park, and Dongwon Lee. 2021.
\newblock \href {https://doi.org/10.18653/v1/2021.acl-long.296} {A sweet rabbit
  hole by {DARCY}: Using honeypots to detect universal trigger{'}s adversarial
  attacks}.
\newblock In \emph{Proceedings of the 59th Annual Meeting of the Association
  for Computational Linguistics and the 11th International Joint Conference on
  Natural Language Processing (Volume 1: Long Papers)}, pages 3831--3844,
  Online. Association for Computational Linguistics.

\bibitem[{Le et~al.(2020)Le, Wang, and Lee}]{insert}
Thai Le, Suhang Wang, and Dongwon Lee. 2020.
\newblock \href {http://arxiv.org/abs/2009.01048} {Malcom: Generating malicious
  comments to attack neural fake news detection models}.

\bibitem[{Li et~al.(2021)Li, Zhang, Peng, Chen, Brockett, Sun, and
  Dolan}]{clare}
Dianqi Li, Yizhe Zhang, Hao Peng, Liqun Chen, Chris Brockett, Ming-Ting Sun,
  and Bill Dolan. 2021.
\newblock \href {http://arxiv.org/abs/2009.07502} {Contextualized perturbation
  for textual adversarial attack}.

\bibitem[{Li et~al.(2019)Li, Ji, Du, Li, and Wang}]{textbugger}
Jinfeng Li, Shouling Ji, Tianyu Du, Bo~Li, and Ting Wang. 2019.
\newblock \href {https://doi.org/10.14722/ndss.2019.23138} {Textbugger:
  Generating adversarial text against real-world applications}.
\newblock \emph{Proceedings 2019 Network and Distributed System Security
  Symposium}.

\bibitem[{Li et~al.(2020)Li, Ma, Guo, Xue, and Qiu}]{BertAttack}
Linyang Li, Ruotian Ma, Qipeng Guo, Xiangyang Xue, and Xipeng Qiu. 2020.
\newblock \href {https://doi.org/10.18653/v1/2020.emnlp-main.500}
  {{BERT}-{ATTACK}: Adversarial attack against {BERT} using {BERT}}.
\newblock In \emph{Proceedings of the 2020 Conference on Empirical Methods in
  Natural Language Processing (EMNLP)}, pages 6193--6202.

\bibitem[{Liu et~al.(2019{\natexlab{a}})Liu, Ott, Goyal, Du, Joshi, Chen, Levy,
  Lewis, Zettlemoyer, and Stoyanov}]{liu2019roberta}
Yinhan Liu, Myle Ott, Naman Goyal, Jingfei Du, Mandar Joshi, Danqi Chen, Omer
  Levy, Mike Lewis, Luke Zettlemoyer, and Veselin Stoyanov. 2019{\natexlab{a}}.
\newblock Roberta: A robustly optimized bert pretraining approach.
\newblock \emph{arXiv preprint arXiv:1907.11692}.

\bibitem[{Liu et~al.(2019{\natexlab{b}})Liu, Liu, Liu, Xu, Lin, Wang, and
  Wen}]{cv-rand1}
Zihao Liu, Qi~Liu, Tao Liu, Nuo Xu, Xue Lin, Yanzhi Wang, and Wujie Wen.
  2019{\natexlab{b}}.
\newblock \href {http://arxiv.org/abs/1803.05787} {Feature distillation:
  Dnn-oriented jpeg compression against adversarial examples}.

\bibitem[{Loshchilov and Hutter(2018)}]{DBLP:journals/corr/abs-1711-05101}
Ilya Loshchilov and Frank Hutter. 2018.
\newblock \href {https://openreview.net/forum?id=rk6qdGgCZ} {Fixing weight
  decay regularization in adam}.

\bibitem[{Maas et~al.(2011)Maas, Daly, Pham, Huang, Ng, and Potts}]{IMDB}
Andrew~L. Maas, Raymond~E. Daly, Peter~T. Pham, Dan Huang, Andrew~Y. Ng, and
  Christopher Potts. 2011.
\newblock \href {https://www.aclweb.org/anthology/P11-1015} {Learning word
  vectors for sentiment analysis}.
\newblock In \emph{Proceedings of the 49th Annual Meeting of the Association
  for Computational Linguistics: Human Language Technologies}, pages 142--150,
  Portland, Oregon, USA. Association for Computational Linguistics.

\bibitem[{Maheshwary et~al.(2020)Maheshwary, Maheshwary, and Pudi}]{HardLabel}
Rishabh Maheshwary, Saket Maheshwary, and Vikram Pudi. 2020.
\newblock Generating natural language attacks in a hard label black box
  setting.

\bibitem[{Malykh(2019)}]{noise1}
Valentin Malykh. 2019.
\newblock \href {https://doi.org/10.18653/v1/P19-2002} {Robust to noise models
  in natural language processing tasks}.
\newblock In \emph{Proceedings of the 57th Annual Meeting of the Association
  for Computational Linguistics: Student Research Workshop}, pages 10--16,
  Florence, Italy. Association for Computational Linguistics.

\bibitem[{Meng and Wattenhofer(2020)}]{white-box1}
Zhao Meng and Roger Wattenhofer. 2020.
\newblock \href {https://doi.org/10.18653/v1/2020.coling-main.585} {A
  geometry-inspired attack for generating natural language adversarial
  examples}.
\newblock In \emph{Proceedings of the 28th International Conference on
  Computational Linguistics}, pages 6679--6689, Barcelona, Spain (Online).
  International Committee on Computational Linguistics.

\bibitem[{Mrk{\v{s}}i{\'c} et~al.(2016)Mrk{\v{s}}i{\'c}, {\'O}~S{\'e}aghdha,
  Thomson, Ga{\v{s}}i{\'c}, Rojas-Barahona, Su, Vandyke, Wen, and
  Young}]{CounterFittingWordVectors}
Nikola Mrk{\v{s}}i{\'c}, Diarmuid {\'O}~S{\'e}aghdha, Blaise Thomson, Milica
  Ga{\v{s}}i{\'c}, Lina~M. Rojas-Barahona, Pei-Hao Su, David Vandyke,
  Tsung-Hsien Wen, and Steve Young. 2016.
\newblock \href {https://doi.org/10.18653/v1/N16-1018} {Counter-fitting word
  vectors to linguistic constraints}.
\newblock In \emph{Proceedings of the 2016 Conference of the North {A}merican
  Chapter of the Association for Computational Linguistics: Human Language
  Technologies}, pages 142--148, San Diego, California. Association for
  Computational Linguistics.

\bibitem[{Naber(2003)}]{languagetool}
D.~Naber. 2003.
\newblock \href {https://books.google.fr/books?id=yKPswAEACAAJ} {\emph{A
  Rule-Based Style and Grammar Checker}}.
\newblock GRIN Verlag.

\bibitem[{Nangia et~al.(2017)Nangia, Williams, Lazaridou, and Bowman}]{MNLI}
Nikita Nangia, Adina Williams, Angeliki Lazaridou, and Samuel~R Bowman. 2017.
\newblock The repeval 2017 shared task: Multi-genre natural language inference
  with sentence representations.
\newblock In \emph{RepEval}.

\bibitem[{Pang and Lee(2005)}]{mr}
Bo~Pang and Lillian Lee. 2005.
\newblock \href {https://doi.org/10.3115/1219840.1219855} {Seeing stars:
  Exploiting class relationships for sentiment categorization with respect to
  rating scales}.
\newblock In \emph{Proceedings of the 43rd Annual Meeting of the Association
  for Computational Linguistics ({ACL}{'}05)}, pages 115--124, Ann Arbor,
  Michigan. Association for Computational Linguistics.

\bibitem[{Pruthi et~al.(2019)Pruthi, Dhingra, and Lipton}]{CombatMisspellings}
Danish Pruthi, Bhuwan Dhingra, and Zachary~C. Lipton. 2019.
\newblock \href {https://doi.org/10.18653/v1/P19-1561} {Combating adversarial
  misspellings with robust word recognition}.
\newblock In \emph{Proceedings of the 57th Annual Meeting of the Association
  for Computational Linguistics}, pages 5582--5591, Florence, Italy.
  Association for Computational Linguistics.

\bibitem[{Raff et~al.(2019)Raff, Sylvester, Forsyth, and McLean}]{cv-rand3}
Edward Raff, Jared Sylvester, Steven Forsyth, and Mark McLean. 2019.
\newblock \href {https://doi.org/10.1109/CVPR.2019.00669} {Barrage of random
  transforms for adversarially robust defense}.
\newblock In \emph{2019 IEEE/CVF Conference on Computer Vision and Pattern
  Recognition (CVPR)}, pages 6521--6530.

\bibitem[{Socher et~al.(2013)Socher, Perelygin, Wu, Chuang, Manning, Ng, and
  Potts}]{SST2}
Richard Socher, Alex Perelygin, Jean Wu, Jason Chuang, Christopher~D. Manning,
  Andrew Ng, and Christopher Potts. 2013.
\newblock \href {https://www.aclweb.org/anthology/D13-1170} {Recursive deep
  models for semantic compositionality over a sentiment treebank}.
\newblock In \emph{Proceedings of the 2013 Conference on Empirical Methods in
  Natural Language Processing}, pages 1631--1642, Seattle, Washington, USA.
  Association for Computational Linguistics.

\bibitem[{Wallace et~al.(2019)Wallace, Feng, Kandpal, Gardner, and
  Singh}]{unitrigger}
Eric Wallace, Shi Feng, Nikhil Kandpal, Matt Gardner, and Sameer Singh. 2019.
\newblock \href {https://doi.org/10.18653/v1/D19-1221} {Universal adversarial
  triggers for attacking and analyzing {NLP}}.
\newblock In \emph{Proceedings of the 2019 Conference on Empirical Methods in
  Natural Language Processing and the 9th International Joint Conference on
  Natural Language Processing (EMNLP-IJCNLP)}, pages 2153--2162, Hong Kong,
  China. Association for Computational Linguistics.

\bibitem[{Wallace et~al.(2020)Wallace, Stern, and Song}]{decision-based1}
Eric Wallace, Mitchell Stern, and Dawn Song. 2020.
\newblock \href {https://doi.org/10.18653/v1/2020.emnlp-main.446} {Imitation
  attacks and defenses for black-box machine translation systems}.
\newblock In \emph{Proceedings of the 2020 Conference on Empirical Methods in
  Natural Language Processing (EMNLP)}, pages 5531--5546, Online. Association
  for Computational Linguistics.

\bibitem[{Wang et~al.(2021)Wang, Liu, Gui, Zhang et~al.}]{textflint}
Xiao Wang, Qin Liu, Tao Gui, Qi~Zhang, et~al. 2021.
\newblock \href {https://doi.org/10.18653/v1/2021.acl-demo.41} {Textflint:
  Unified multilingual robustness evaluation toolkit for natural language
  processing}.
\newblock In \emph{Proceedings of the 59th Annual Meeting of the Association
  for Computational Linguistics and the 11th International Joint Conference on
  Natural Language Processing: System Demonstrations}, pages 347--355, Online.
  Association for Computational Linguistics.

\bibitem[{Wei and Zou(2019)}]{eda}
Jason Wei and Kai Zou. 2019.
\newblock \href {https://doi.org/10.18653/v1/D19-1670} {{EDA}: Easy data
  augmentation techniques for boosting performance on text classification
  tasks}.
\newblock In \emph{Proceedings of the 2019 Conference on Empirical Methods in
  Natural Language Processing and the 9th International Joint Conference on
  Natural Language Processing (EMNLP-IJCNLP)}, pages 6382--6388, Hong Kong,
  China. Association for Computational Linguistics.

\bibitem[{Ye et~al.(2020)Ye, Gong, and Liu}]{safer}
Mao Ye, Chengyue Gong, and Qiang Liu. 2020.
\newblock \href {https://doi.org/10.18653/v1/2020.acl-main.317} {{SAFER}: A
  structure-free approach for certified robustness to adversarial word
  substitutions}.
\newblock In \emph{Proceedings of the 58th Annual Meeting of the Association
  for Computational Linguistics}, pages 3465--3475, Online. Association for
  Computational Linguistics.

\bibitem[{Zhang et~al.(2020)Zhang, Sheng, Alhazmi, and Li}]{survey1}
Wei~Emma Zhang, Quan~Z. Sheng, Ahoud Alhazmi, and Chenliang Li. 2020.
\newblock \href {https://doi.org/10.1145/3374217} {Adversarial attacks on
  deep-learning models in natural language processing: A survey}.
\newblock \emph{ACM Trans. Intell. Syst. Technol.}, 11(3).

\bibitem[{Zhao et~al.(2018)Zhao, Dua, and Singh}]{gan}
Zhengli Zhao, Dheeru Dua, and Sameer Singh. 2018.
\newblock Generating natural adversarial examples.
\newblock In \emph{International Conference on Learning Representations
  (ICLR)}.

\bibitem[{Zhou et~al.(2019)Zhou, Jiang, Chang, and Wang}]{DISP}
Yichao Zhou, Jyun{-}Yu Jiang, Kai{-}Wei Chang, and Wei Wang. 2019.
\newblock Learning to discriminate perturbations for blocking adversarial
  attacks in text classification.
\newblock In \emph{Proceedings of the 2019 Conference on Empirical Methods in
  Natural Language Processing and the 9th International Joint Conference on
  Natural Language Processing, {EMNLP-IJCNLP} 2019, Hong Kong, China, November
  3-7, 2019}. Association for Computational Linguistics.

\end{thebibliography}
\bibliographystyle{acl_natbib}

\end{document}